\documentclass{nature_meth}
\usepackage{amssymb,amsfonts,amsmath}
\usepackage{graphicx} 
\usepackage{cite}
\usepackage{subfig}
\usepackage{amsmath,amssymb,amsfonts}
\usepackage{algorithmic}
\usepackage{graphicx}
\usepackage{hyperref}
\usepackage{multirow}
\usepackage{multicol}
\usepackage{booktabs}
\usepackage{makecell}
\usepackage{tabularx}
\usepackage{gensymb}
\hypersetup{colorlinks=true,urlcolor=red}
\usepackage{wrapfig}
\usepackage{multirow}
\usepackage{epsfig}
\usepackage{hyperref}
\usepackage{xcolor}
\hypersetup{
    colorlinks=true,
    linkcolor=blue,
    filecolor=magenta,
    urlcolor=cyan,
}

\newcommand{\yi}[1] {\textcolor{black}{#1}}

\usepackage{caption,setspace}
\renewcommand{\spacing}[1]{\renewcommand{\baselinestretch}{#1}\large\normalsize}
\spacing{2}
\usepackage{url}
\usepackage{lineno}
\usepackage{booktabs}
\usepackage{adjustbox}
\usepackage{marginnote}
\renewcommand{\marginnote}[2][]{}

\usepackage{graphicx}
\makeatletter
\let\saved@includegraphics\includegraphics
\AtBeginDocument{\let\includegraphics\saved@includegraphics}
\makeatother

\usepackage{kotex}
\usepackage{tablefootnote}

\title{RadFabric: Agentic AI System with Reasoning Capability for Radiology}

\author{Wenting Chen$^{1,2\ast}$, 
Yi Dong$^{3\ast}$, 
Zhaojun Ding$^{4}$,
Yucheng Shi$^{4}$,
Yifan Zhou$^{4}$,
Fang Zeng$^{2}$,
Yijun Luo$^{2}$,\\
Tianyu Lin$^{2}$,
Yihang Su$^{3}$,
Yichen Wu$^{2}$,
Kai Zhang$^{5}$,
Zhen Xiang$^{4}$,
Tianming Liu$^{4}$,
Ninghao Liu$^{4}$,\\
Lichao Sun$^{5\dagger}$,
Yixuan Yuan$^{1,3\dagger}$, and
Xiang Li$^{2\dagger}$
}

\begin{document}
\setstretch{1}
\maketitle

\begin{affiliations}
\item{Department of Electrical Engineering, City University of Hong Kong, Hong Kong SAR, China}
\item{Department of Radiology, Massachusetts General Hospital, Boston, MA, USA}
\item{Department of Electronic Engineering, The Chinese University of Hong Kong, Hong Kong SAR, China}
\item{School of Computing, University of Georgia, Athens, GA, USA}
\item{Department of Computer Science and Engineering, Lehigh University, Bethlehem, PA, USA}

\item[] $^{\ast}$These authors contributed equally
\item[] $^{\dagger}$Correspondence to Xiang Li (xli60@mgh.harvard.edu), Lichao Sun (lis221@lehigh.edu), and Yixuan Yuan (yxyuan@ee.cuhk.edu.hk)
\end{affiliations}

\section*{Abstract}
\begin{abstract}
Chest X-ray (CXR) imaging remains a critical diagnostic tool for thoracic conditions, but current automated systems face limitations in pathology coverage, diagnostic accuracy, and integration of visual and textual reasoning. To address these gaps, we propose RadFabric, a multi-agent, multimodal reasoning framework that unifies visual and textual analysis for comprehensive CXR interpretation. RadFabric is built on the Model Context Protocol (MCP), enabling modularity, interoperability, and scalability for seamless integration of new diagnostic agents. The system employs specialized CXR agents for pathology detection, an Anatomical Interpretation Agent to map visual findings to precise anatomical structures, and a Reasoning Agent powered by large multimodal reasoning models to synthesize visual, anatomical, and clinical data into transparent and evidence-based diagnoses. RadFabric achieves significant performance improvements, with near-perfect detection of challenging pathologies like fractures (1.000 accuracy) and superior overall diagnostic accuracy (0.799) compared to traditional systems (0.229–0.527). By integrating cross-modal feature alignment and preference-driven reasoning, RadFabric advances AI-driven radiology toward transparent, anatomically precise, and clinically actionable CXR analysis.

\end{abstract}
 
\clearpage

\section*{Introduction}
\label{sec:intro}



Chest X-ray (CXR) imaging remains a cornerstone of thoracic diagnostics, enabling rapid detection of critical conditions such as pneumonia, pneumothorax, and cardiomegaly. \yi{Despite its ubiquity, clinical interpretation of CXRs still largely relies on manual reading by radiologists, which is subject to inter-observer variability, time constraints, and the growing volume of imaging studies.} These limitations underscore the urgent need for effective foundation models \yi{equipped with high explainability and context-aware reasoning, capable of enhancing clinical decision-making with greater speed, accuracy, and transparency.}

\yi{While recent advances in automated CXR foundation models show promise, existing approaches still fall short in several critical areas: \textcircled{1} Narrow Pathology Coverage. Most existing foundation models have narrow pathology coverage and act as specialized expert systems, often showing inconsistent performance across different pathologies. For instance, they often excel in detecting pathologies like pleural effusion (with performance up to 0.783), but fail to generalize to others such as enlarged cardiomediastinum or lung lesions. \textcircled{2} Limited Clinical Applicability. Clinical applicability is often limited by inadequate integration of visual information and a lack of interactivity with clinical environments.
Some foundation models are unable to integrate visual information such as lesion localization or anatomical context with clinical reasoning, which limits their diagnostic effectiveness in complex cases. Although recent advances in multi-modal large language models (LLMs) show promise in lesion detection and report generation, these models remain disconnected from real-world clinical workflows. They cannot interact with external systems, revise their reasoning based on new evidence, or incorporate contextual information. This lack of interactivity limits their practical use in clinical settings, where adaptability and real-time decision-making are critical. To address these aforementioned two limitations, one promising solution lies in leveraging the complementary strengths of small vision models and large language models. Small vision models have usually demonstrated strong performance in pathology and lesion detection, recognition, and classification tasks, proving effective in specialized visual domains. In contrast, although LLMs often fall short in visual pathology detection, they offer advanced reasoning and contextual understanding.  Combining their complementary strengths can improve visual integration and enable dynamic interactivity, supporting effective interaction with clinical environments, continuous updating based on new evidence, and real-time incorporation of contextual information.}  
\yi{Recent advances in large reasoning models (LRMs), such as DeepSeek-R1 and OpenAI-O1, have demonstrated strong capabilities in reasoning and contextual understanding, highlighting their strong potential for applications in medical AI. Specifically,} these models excel at synthesizing multi-source textual data, resolving contradictions, and generating logically coherent conclusions. However, their reliance on text-only paradigms \yi{restricts their capacity to interpret visual information, which is essential in radiology and directly informs clinical decisions.} For instance, distinguishing pneumonia from atelectasis requires not only detecting lung opacity but also correlating its spatial distribution with clinical indicators. \yi{Therefore, how to bridge the gap and develop a unified framework that integrates both visual and textual reasoning has emerged as a key challenge. 
A deeper challenge involves converting visual findings into anatomically accurate clinical descriptions, which demands both detailed visual understanding and clinical expertise.
Overcoming this requires a flexible framework that enables dynamic interaction between the reasoning agent, the environment, and the data. Such rich interaction enables the model not only to interpret findings accurately, but also to iteratively refine its understanding through continuous engagement with both contextual and visual information.}
\begin{figure}[t]
  \centering
   \includegraphics[width=\linewidth]{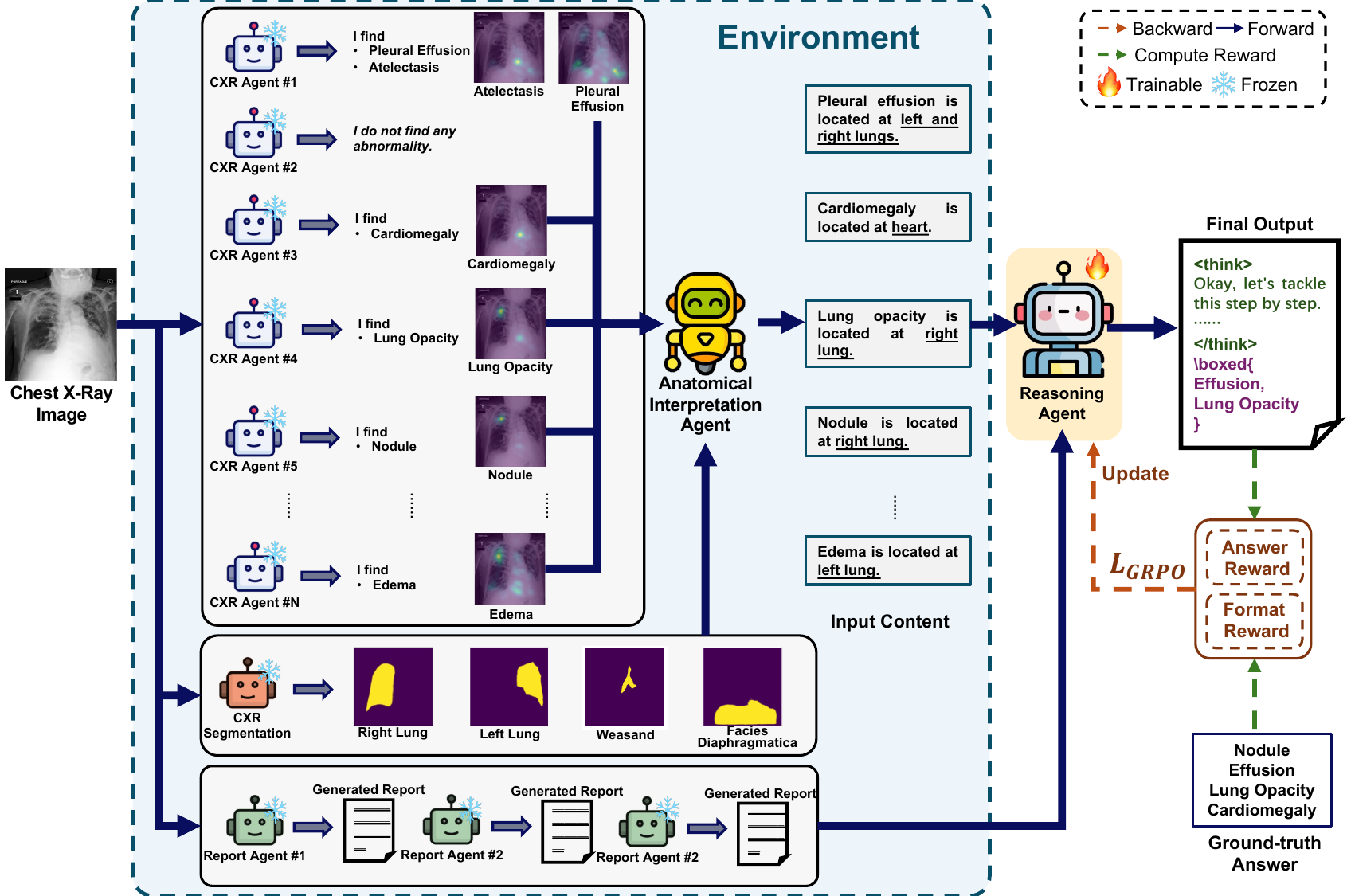}
   \caption{Illustrative diagram of the RadFabric framework}
   \label{main_fig}
\end{figure}

\yi{To address these challenges, we introduce RadFabric,} a multimodal reasoning framework that unifies visual and textual reasoning and supports effective interaction with clinical environments for a comprehensive interpretation of CXR. \yi{As shown in Fig.~\ref{main_fig}, the proposed RadFabric contains four parts: The \textit{CXR Agent}, which} employs small, highly effective vision models for precise pathology detection and generates interpretable Grad-CAM maps that highlight regions of interest (e.g., fracture sites, pleural effusion). These specialized models excel in detecting and localizing pathologies with accuracy, addressing the limitations of large language models in direct visual analysis. \yi{The \textit{Anatomical Interpretation Agent}, which} anchors these visual findings to segmented anatomical structures (e.g., left lung, diaphragm), transforming heatmaps into precise clinical descriptions (e.g., “effusion localized to the left costophrenic angle”). By integrating these specialized vision models, RadFabric significantly enhances the diagnostic performance of the overall system. \yi{The \textit{Report Agent}, which} utilizes multimodal models (e.g., Qwen2-VL-7b) to generate structured clinical reports. \yi{And the \textit{Reasoning Agent}, which integrates visual maps, anatomical context, and textual reports,} is interactive, and explicitly trainable to produce step-by-step reasoning before generating the final diagnosis. This process enhances interpretability, as the reasoning trajectory itself provides a transparent, clinically meaningful rationale for each diagnosis.
\yi{The proposed RadFabric integrates visual and textual reasoning through a modular, multi-agent architecture that enables dynamic interaction with both data and environment. By decoupling the roles of model, data, and environment, it promotes flexibility and scalability. Specialized agents, such as lightweight CXR Agent, serve as tools for a central reasoning agent, and can be independently updated to enhance performance over time. This design allows the system to iteratively refine its understanding, leading to more accurate, interpretable, and clinically grounded decisions.} {Empirically, RadFabric achieves near-perfect fracture detection (1.000 vs. 0.096–0.269 in legacy systems) and significantly improves lung lesion identification (0.850 vs. 0.176–0.197), setting a new standard for reliable and actionable CXR interpretation. }

This study presents the development, validation, and clinical evaluation of RadFabric. Subsequent sections detail its methodology, benchmark performance, and implications for AI-driven radiology. By unifying visual and textual reasoning—and integrating specialized models to enhance performance—RadFabric demonstrates the potential for robust, multimodal diagnostic systems in medical imaging.

\begin{table}
    \centering
    \renewcommand{\arraystretch}{1.2}  
    \setlength\tabcolsep{2pt}
    \caption{Chest X-Ray (CXR) Agents and their pathology coverage (Acc).}
    \label{cxr_res_part1}
    \small
    \scalebox{0.87}{\begin{tabular}{ccccccccc}
    \bottomrule[1pt] \hline 
    Agents & Atelectasis & Cardiomegaly & Consolidation & Edema & \makecell{Enlarged\\Cardiomediastinum} & Fracture & \makecell{Lung\\Lesion} & \\ \hline
    CXR Agent 1 & 0.195 & 0.232 & 0.203 & 0.312 & 0.138 & 0.096 & 0.176  & \\
    CXR Agent 2 & 0.212 & 0.292 & 0.287 & 0.414 & 0.134 & 0.269 & 0.197  & \\
    CXR Agent 3 & - & & - & & - & & -  & \\
    CXR Agent 4 & 0.355 & 0.572 & 0.121 & 0.423 & - & & - &  \\
    CXR Agent 5 & 0.253 & 0.384 & 0.414 & 0.609 & - & & -  & \\
    CXR Agent 6 & 0.655 & 0.383 & 0.435 & 0.378 & - & & -  & \\
    CXR Agent 7 & 0.234 & 0.348 & 0.256 & 0.331 & - & & -  & \\ \hline
    RadFabric-o1 & 0.889 & 0.722 & 0.842 & 0.833 & 1.000 & 1.000 & 0.850  & \\
    RadFabric-R1 & 0.611 & 0.722 & 0.737 & 0.667 & 0.950 & 0.950 & 0.650  & \\
    \bottomrule[1pt] \hline 
    Agents & \makecell{Lung\\Opacity} & \makecell{No\\Finding} & \makecell{Pleural\\Effusion} & \makecell{Pleural\\Other} & Pneumonia & Pneumothorax & \makecell{Support\\Devices} & Overall \\ \hline
    CXR Agent 1 & 0.355 & - & 0.375 & - & 0.213 & 0.218 & - & 0.229 \\
    CXR Agent 2 & 0.355 & - & 0.428 & - & 0.246 & 0.215 & - & 0.277 \\
    CXR Agent 3 & 0.513 & - & & - & 0.469 & - & & 0.491 \\
    CXR Agent 4 & & - & 0.486 & - & 0.234 & 0.411 & - & 0.372 \\
    CXR Agent 5 & & - & 0.527 & - & 0.348 & 0.447 & - & 0.426 \\
    CXR Agent 6 & & - & 0.783 & - & & - & & 0.527 \\
    CXR Agent 7 & & - & 0.446 & - & & - & & 0.323 \\ \hline
    RadFabric-o1 & 0.700 & 0.700 & 0.400 & 0.850 & 0.867 & 0.850 & 0.700 & 0.799 \\
    RadFabric-R1 & 0.650 & 0.750 & 0.450 & 0.900 & 0.733 & 0.850 & 0.700 & 0.739 \\
    \bottomrule[1pt] \hline
    \end{tabular}}
\end{table}

\section*{Results of RadFabric with Frozen Reasoning Agent}
Traditional CXR Agents (1-7) demonstrate modest diagnostic capabilities with overall performance scores ranging from 0.229 to 0.527, and substantial variability across different pathologies. These agents generally perform better on conditions like Pleural Effusion (0.375-0.783) and Edema (0.312-0.609), while struggling with Fracture detection (0.096-0.269) and Lung Lesion identification (0.176-0.197). Notably, most traditional agents exhibit significant coverage gaps, with many unable to detect certain pathologies entirely, suggesting specialized rather than comprehensive diagnostic utility.

In contrast, the novel RadFabric agents (RadFabric-o1 and RadFabric-R1) represent a significant advancement with overall performance scores of 0.799 and 0.739 respectively, substantially outperforming all traditional counterparts. These agents provide comprehensive coverage across all 14 pathologies, with RadFabric-o1 achieving perfect scores (1.000) for Enlarged Cardiomediastinum and Fracture detection—conditions where traditional agents either perform poorly or lack capabilities entirely. This marked improvement in both performance and pathology coverage suggests that newer RadFabric technology offers considerably more reliable and versatile diagnostic support for clinical chest X-ray interpretation.

\begin{figure}[t]
  \centering
   \includegraphics[width=\linewidth]{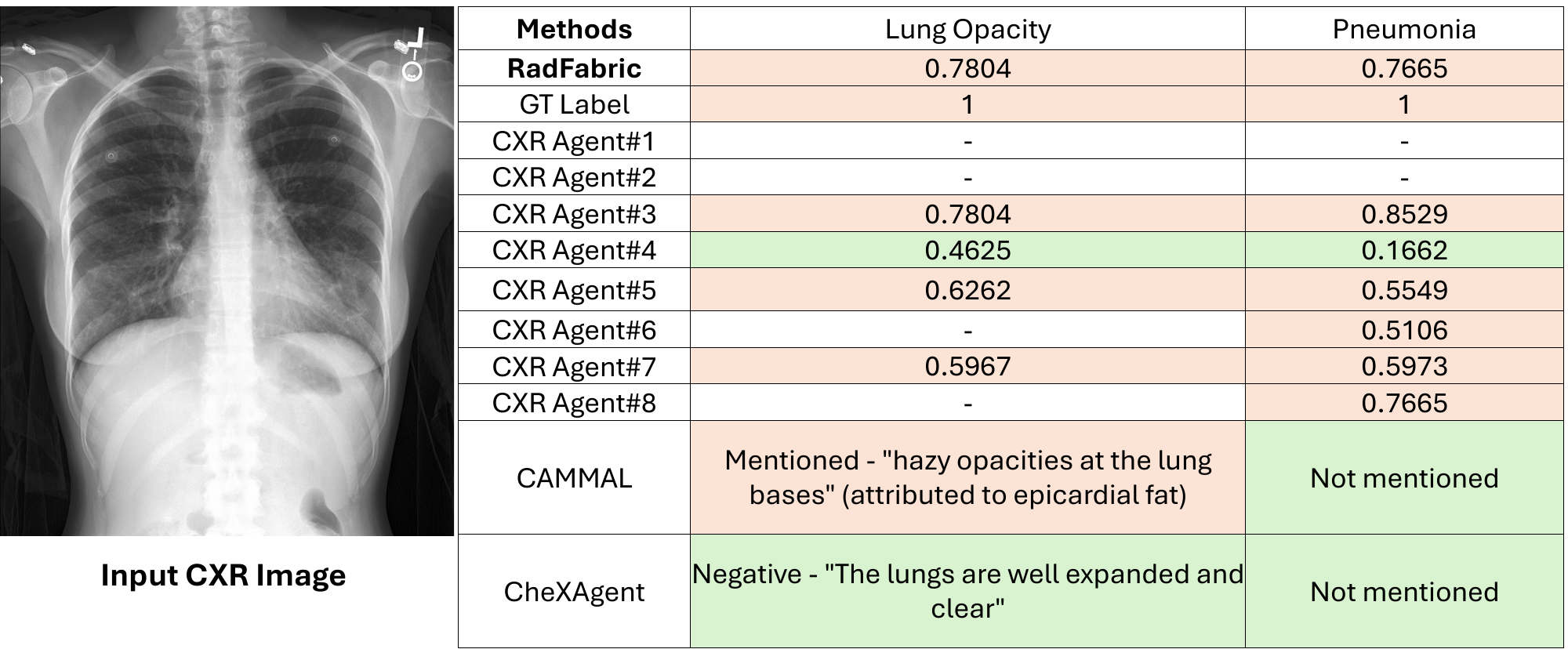}
   \caption{Comparative results from our RadFabric system, CXR agents, and established report generation methods (CAMMAL and CheXagent) for chest x-ray image 1.}
   \label{case1}
\end{figure}

In Table~\ref{case1}, the results demonstrate the strengths and limitations of different methods in identifying lung opacity and pneumonia in chest X-rays. While the CXR agent (classification model) shows strong predictive capabilities for lesions, as seen in cases such as lung opacity (e.g., CXR Agent\#3: 0.7804) and pneumonia (e.g., CXR Agent\#3: 0.8529), report generation models like CAMMAL and CheXAgent may occasionally fail to explicitly mention these findings. For instance, CAMMAL noted "hazy opacities at the lung bases" but attributed them to epicardial fat, while CheXAgent reported negative findings for the lungs. This highlights the potential for complementary use, where the classification model can detect lesions that report generators might overlook. RadFabric, our proposed method, integrates multiple CXR agents and report generation models, enabling a more robust analysis. By leveraging diverse perspectives, RadFabric minimizes the likelihood of missing lesions, achieving predictions that closely align with the ground truth labels (e.g., lung opacity: 0.7804, pneumonia: 0.7665). This integration underscores its potential for improving diagnostic accuracy.

\begin{figure}[t]
  \centering
   \includegraphics[width=\linewidth]{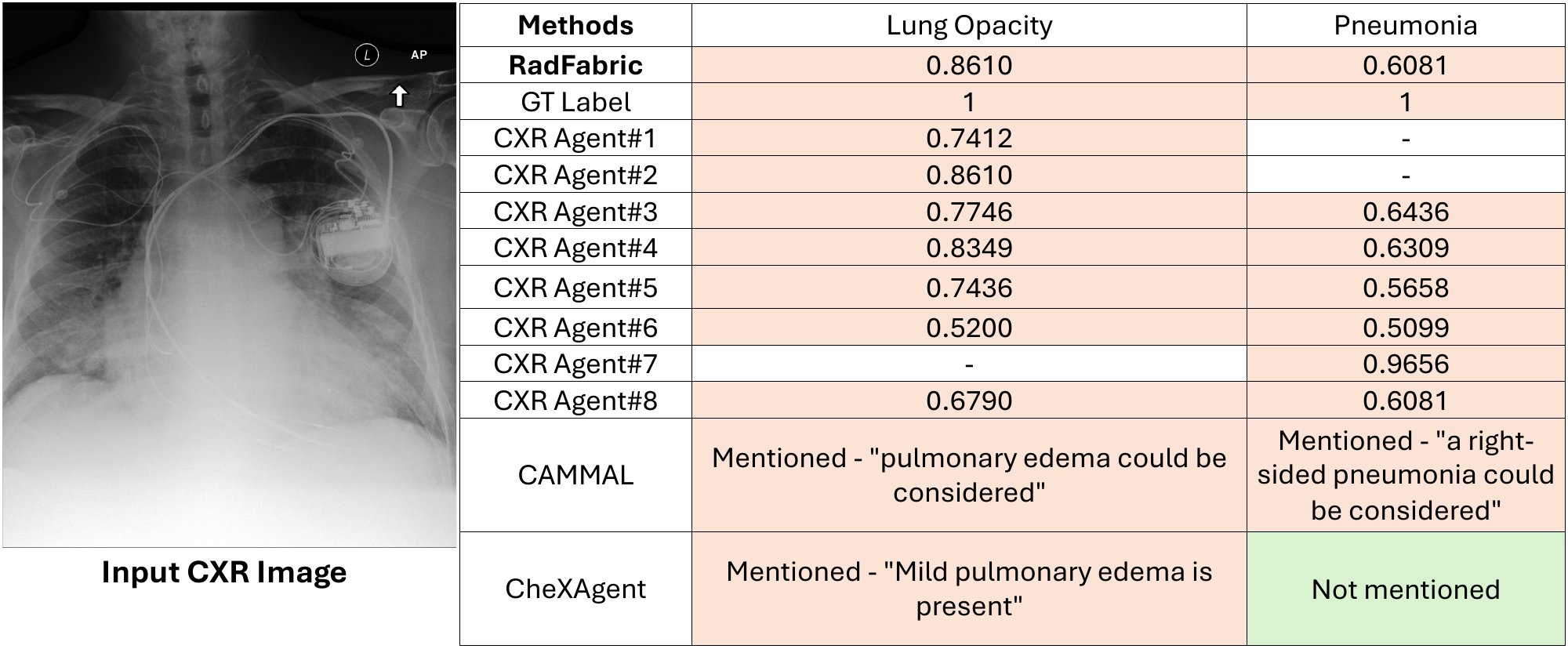}
   \caption{Visual Results of our RadFabric, CXR agents, and two report generation methods (i.e. CAMMAL and CheXagent) for chest x-ray image 2.}
   \label{case2}
\end{figure}

As displayed in Fig.~\ref{case2}, the results also highlight the variability and potential biases of individual models, emphasizing the importance of integrating multiple perspectives. For example, while CXR Agent\#2 and CXR Agent\#3 provided high scores for lung opacity (0.861 and 0.7746, respectively), they differed significantly in their pneumonia predictions, with CXR Agent\#2 failing to provide a score and CXR Agent\#3 predicting 0.6436. Similarly, CXR Agent\#7 showed a high pneumonia prediction (0.9656) but did not provide a lung opacity score. This inconsistency across models indicates that relying on a single agent may lead to incomplete or skewed results. Additionally, the report generation models, CAMMAL and CheXAgent, not only differ in the findings they report but also in how they interpret the clinical context. CAMMAL mentioned both "pulmonary edema" and "a right-sided pneumonia," showing a broader diagnostic scope, while CheXAgent focused on "mild pulmonary edema" and omitted any mention of pneumonia. This discrepancy indicates that report generation models are subject to interpretive limitations and may miss critical findings depending on the phrasing or contextual emphasis.

RadFabric addresses these challenges by combining the strengths of both classification models and report generators. Its ability to aggregate and reconcile outputs ensures a more balanced and complete understanding of potential abnormalities. For instance, RadFabric captures the high lung opacity score (0.861) from key agents like CXR Agent\#2 while maintaining sensitivity to pneumonia findings (0.6081) by incorporating information from agents and contextual cues from report generators. This multi-faceted approach reduces reliance on any single model's performance and mitigates the risk of diagnostic gaps, making RadFabric a more reliable and robust solution for clinical applications.

\section*{Results of RadFabric with Trainable Reasoning Agent}
When we train the reasoning agent with the GRPO strategy, the overall accuracy achieves 0.897, which surpasses the RadFabric with the frozen reasoning agent by a large margin of 9.8\%. This suggests the trainable reasoning agent can learn and adapt to the nuanced requirements of clinical diagnosis, improving its ability to synthesize multimodal data for more precise and contextually relevant conclusions. The Guided Reward Policy Optimization (GRPO) strategy further enhances the agent's capacity to prioritize clinically significant reasoning pathways, ensuring that the generated diagnoses are both evidence-based and aligned with real-world medical expectations. This adaptability is particularly significant in complex cases where subtle or overlapping pathologies may otherwise be misinterpreted or overlooked. For instance, the trainable reasoning agent demonstrates superior performance in distinguishing conditions with similar visual manifestations, such as pneumonia versus atelectasis. By leveraging cross-modal attention mechanisms and iterative learning, the agent refines its understanding of spatial patterns in visual data (e.g., Grad-CAM heatmaps) and correlates them with textual inputs like patient history or symptom descriptions. This capability not only improves diagnostic accuracy but also enhances transparency, as the reasoning process can be traced back to specific visual and textual evidence.

\begin{figure}[t]
  \centering
   \includegraphics[width=\linewidth]{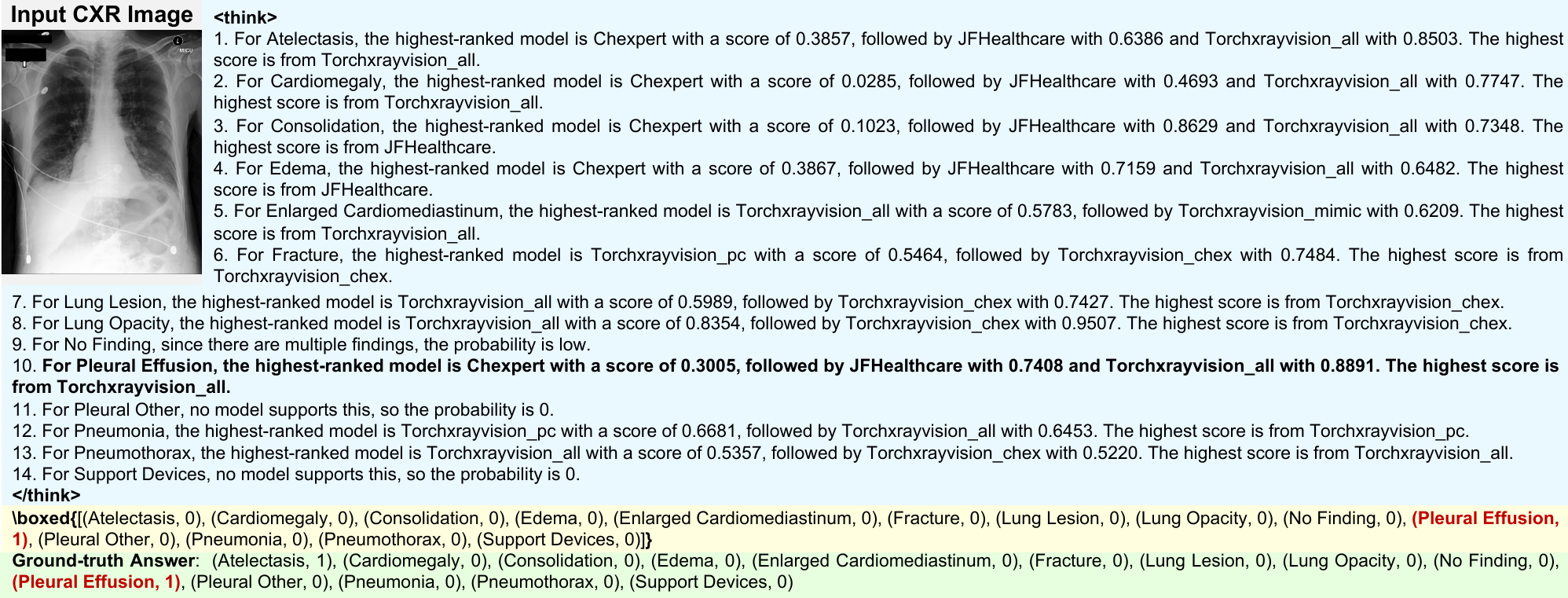}
   \caption{Visual Results of our RadFabric with trainable reasoning agent.}
   \label{case_reason}
\end{figure}

In Fig.~\ref{case_reason}, the visual result generated by the RadFabric system with a trainable reasoning agent demonstrates both the strengths and current limitations of multi-agent, multimodal CXR analysis. As shown in the <think> reasoning trace, the system leverages multiple specialized models—such as Chexpert, JFHealthcare, and various Torchxrayvision variants—to independently assess a wide spectrum of pathologies. For most conditions, the model correctly identifies the absence of findings, and it successfully detects pleural effusion, in agreement with the ground-truth label. Notably, the agent assigns the highest probability for pleural effusion based on the Torchxrayvision\_all model, which aligns with the reference standard. However, the system fails to recognize the presence of atelectasis, despite high scores from several component models (e.g., Torchxrayvision\_all: 0.8503), ultimately outputting a negative prediction for this pathology. This discrepancy highlights a challenge in model aggregation and decision fusion, where high individual model confidence does not always translate into a positive final prediction—potentially due to conservative thresholding or conflicting evidence among agents. The visual evidence, likely reflected in the Grad-CAM heatmaps, supports the model’s high confidence for pleural effusion, suggesting robust localization and anatomical grounding for this finding. Overall, the result exemplifies RadFabric’s ability to synthesize multi-source data and generate interpretable outputs, yet also underscores the importance of further optimizing integration strategies to reduce false negatives, particularly in cases of co-existing pathologies such as atelectasis.

\section*{Method}
\subsection{Overview}
\yi{To address the need for faster, more accurate, and transparent chest X-ray (CXR) diagnosis, we developed RadFabric. This multi-agent system functions as an explainable, context-aware foundation model that integrates visual analysis with clinical reasoning to assist or automate radiological interpretation. The analytical workflow of RadFabric is managed by our distinct agents. The process begins with two parallel inputs: 1) the \textit{CXR Agent Group} provides an initial diagnosis and visual map of potential disease areas, and 2) the \textit{Report Agent Group} creates a structured clinical report. These outputs are then processed by the \textit{Anatomical Interpretation Agent}, which analyzes the spatial location of the visual findings and translates them into precise anatomical terminology. In the final stage, the \textit{Reasoning Agent} integrates all preceding information, including the diagnosis, report, and anatomical analysis, to produce a comprehensive assessment of chest pathologies through higher-order synthesis. The following subsections describe the four components in detail.}

\subsection{CXR Agent Group}
\yi{The \textit{CXR Agent Group} consists of eight specialized agents~\cite{dai2024unichest,cohen2022torchxrayvision,banik2024robust,rajpurkar2017chexnet}, each trained on distinct datasets to detect specific pathologies, as detailed in Table~\ref{cxr_agents}. When presented with a CXR image, each agent independently performs two critical functions. First, it generates a textual diagnostic hypothesis (e.g., "cardiomegaly" or "atelectasis"). Second, it produces a corresponding visual interpretation map using Gradient-weighted Class Activation Mapping (Grad-CAM)~\cite{selvaraju2017grad} to localize the image regions that informed its finding. This dual-output design provides both a clinical assessment and its visual evidence.}

\yi{The collective outputs from all agents--the set of textual hypotheses and their associated visual maps--are then aggregated. This parallelized analysis ensures comprehensive coverage across a wide range of chest abnormalities, with deliberate overlap between agents enhancing detection robustness. This aggregated, multimodal information is then forwarded to the next stage of the RadFabric pipeline, establishing a rich foundation of text and visual evidence for higher-order diagnostic reasoning.}

\subsection{Report Agent Group}
The \textit{Report Agent Group} employs two specialized multimodal models—\yi{ChexAgent}~\cite{chen2024chexagent} and Qwen2-VL-7b~\cite{Qwen2.5-VL}—to generate comprehensive clinical reports from chest radiographs. \yi{This dual-agent approach is a deliberate design choice to enhance the system's robustness and interpretive depth. By having each VLM independently analyze the image, the system benefits from complementary perspectives, as the models may highlight different abnormalities or interpret the same findings with varying clinical emphasis.}

\yi{For a given input image, each agent produces a detailed clinical report that documents relevant observations, potential pathologies, and a preliminary interpretation. These two reports are then aggregated and passed to the subsequent stages of the RadFabric pipeline. There, they serve as critical narrative inputs for the final diagnostic synthesis, ensuring the system's ultimate assessment is informed by both the comprehensive radiological reporting from this group and the targeted pathology detection described in the preceding section. }

\subsection{Anatomical Interpretation Agent}
\yi{The \textit{Anatomical Interpretation Agent} contextualizes visual findings by mapping highlighted disease regions from the CXR Agent Group to their precise locations within the chest radiograph. This process anchors the abstract visual markers to a standardized anatomical framework, thereby enhancing their diagnostic value.}


Specifically, given a CXR image, the agent first performs anatomical segmentation, dividing the radiograph into key structural regions including the esophagus, left lung, right lung, and diaphragmatic surfaces. This segmentation establishes a standardized anatomical framework that serves as a reference map for subsequent analysis. The agent then employs a spatial correlation algorithm to analyze the relationship between GradCAM-highlighted regions from the CXR Agent Group and the segmented anatomical structures. This analysis quantifies the degree of overlap and spatial positioning of potential pathological areas relative to specific anatomical landmarks.

Based on these spatial correlations, the agent generates precise anatomical descriptions in clinical language. For example, \yi{if Grad-CAM highlights indicating “pleural effusion” predominantly overlap with the left lung segment, the agent produces the statement: “The effusion is localized to the left lower lung field, with associated blunting of the costophrenic angle.” This step effectively translates the visual evidence into clinically meaningful spatial information.}
By providing this anatomical precision, the agent enhances the interpretability of the visual findings and facilitates more accurate clinical reasoning in the subsequent stages of the diagnostic pipeline. This anatomical grounding is particularly valuable for conditions where location significantly influences differential diagnosis and treatment planning.


\subsection{Reasoning Agent}
The \textbf{Reasoning Agent} represents the culmination of our RadFabric system, synthesizing inputs from all previous agents to perform sophisticated clinical reasoning and generate comprehensive diagnostic assessments. This agent integrates initial diagnosis results, anatomical context, and preliminary clinical interpretations into a cohesive diagnostic framework.

We leverage advanced large reasoning models—specifically OpenAI o1~\cite{jaech2024openai} or DeepSeek-R1~\cite{ds2025r1}—as the foundation for our reasoning agent due to their exceptional capabilities in complex logical inference, medical knowledge integration, and clinical decision-making. These models have demonstrated superior performance in connecting disparate pieces of evidence and resolving potential contradictions between different information sources. In addition to OpenAI o1 and DeepSeek-R1, we can use other open-source multi-modal large language models (MLLMs) as our reasoning model.

The reasoning process follows a structured multi-step approach: First, the agent aggregates inputs from all preceding components, including initial diagnosis results from the CXR Agent Group, anatomical correlations from the Anatomical Interpretation Agent, and structured reports from the Report Agent Group. This aggregation creates a comprehensive information package with textual evidence. Second, the agent conducts a systematic cross-validation of findings across initial diagnosis results from different CXR agents, identifying consistencies and resolving apparent contradictions. Finally, the agent generates a comprehensive assessment for each potential pathology, assigning confidence levels based on the strength and consistency of supporting evidence. 

To enable the reasoning agent to develop robust and interpretable reasoning capabilities, we employ \textbf{Guided Reward Policy Optimization (GRPO)} as our core training strategy. Under this approach, the model is trained using structured prompts that encourage a "think-then-answer" reasoning pattern: the agent first explicitly articulates its reasoning process—enclosed within delimiters for clarity—and then presents its final diagnostic conclusions in a standardized, easily extractable format. GRPO provides reward signals that incentivize both accurate predictions and adherence to the required format, promoting not only diagnostic correctness but also transparency in the agent's logical pathway. This explicit separation of reasoning and conclusion ensures that each diagnostic output is accompanied by a clear, step-by-step explanation, facilitating interpretability and enabling thorough downstream evaluation of both format adherence and clinical accuracy.



This reasoning-centric approach enhances diagnostic transparency and enables explainable AI by providing clinicians with not only the final diagnostic conclusions but also the logical pathway through which these conclusions were reached. By maintaining a clear chain of evidence from visual findings to anatomical context to clinical reasoning, the RadFabric system offers interpretable and clinically sound chest X-ray analysis that can supplement radiological expertise in clinical settings.



\begin{table}
    \centering
    \renewcommand{\arraystretch}{1.3}  
   \setlength\tabcolsep{1pt}
    \caption{Chest X-Ray (CXR) Agents and their pathology coverage.}
    \label{cxr_agents}
    \small
    \begin{tabular}{c|c|c|c|c|c|c|c}
    \bottomrule[1pt] \hline
    
    \textbf{Pathology} & \textbf{Agent 1} & \textbf{Agent 2} & \textbf{Agent 3} & \textbf{Agent 4} & \textbf{Agent 5}& \textbf{Agent 6}& \textbf{Agent 7} \\ 
    \hline
    Dataset & All Datasets & MIMIC-CXR & \makecell{RSNA \\Pneumonia \\Challenge} & \makecell{NIH Chest \\X-Ray8} & PadChest & JFhealthcare & CheXpert \\ 
    \bottomrule[1pt] \hline
    Atelectasis & \checkmark & \checkmark &  & \checkmark & \checkmark & \checkmark & \checkmark \\ 
    Cardiomegaly & \checkmark & \checkmark &  & \checkmark & \checkmark & \checkmark & \checkmark \\ 
    Consolidation & \checkmark & \checkmark &  & \checkmark & \checkmark & \checkmark & \checkmark \\ 
    Edema & \checkmark & \checkmark &  & \checkmark & \checkmark & \checkmark & \checkmark \\ 
    Effusion & \checkmark & \checkmark &  &  & \checkmark & \checkmark & \checkmark \\ 
    Emphysema & \checkmark &  &  & \checkmark & \checkmark &  &  \\ 
    \makecell{Enlarged\\Cardiomediastinum} & \checkmark & \checkmark &  &  &  &  &  \\ 
    Fracture & \checkmark & \checkmark &  &  & \checkmark &  &  \\ 
    Fibrosis & \checkmark &  &  & \checkmark & \checkmark &  &  \\ 
    Hernia & \checkmark &  &  & \checkmark & \checkmark &  &  \\ 
    Infiltration & \checkmark &  &  &  & \checkmark &  &  \\ 
    Lung Lesion & \checkmark & \checkmark &  &  &  &  &  \\ 
    Lung Opacity & \checkmark & \checkmark & \checkmark &  &  &  &  \\ 
    Mass & \checkmark &  &  & \checkmark & \checkmark &  &  \\ 
    Nodule & \checkmark &  &  & \checkmark & \checkmark &  &  \\ 
    Pleural Other &  &  &  &  &  &  &  \\ 
    Pleural Thickening & \checkmark &  &  & \checkmark & \checkmark &  &  \\ 
    Pneumonia & \checkmark & \checkmark & \checkmark & \checkmark & \checkmark &  &  \\ 
    Pneumothorax & \checkmark & \checkmark &  & \checkmark & \checkmark &  &  \\ 
    \bottomrule[1pt]
    \hline
    \end{tabular}
\end{table}

\subsection{Implementation Details}
The RadFabric framework is implemented on the MCP server, which utilizes the MCP protocol to communicate with various MCP servers. All components of the framework—including CXR agents, report agents, the anatomical interpretation agent, and the reasoning agent—are developed and deployed on the MCP server. An MCP client is designed to interact with the server, enabling the processing of chest X-ray images and generating diagnostic predictions. The reasoning agent in our RadFabric system is trained using a reinforcement learning approach built on the EasyR1 framework, employing Generative Reward-conditioned Policy Optimization (GRPO) to enhance both diagnostic accuracy and interpretability. The base model, Qwen2.5-14B-Instruct, is fine-tuned for chest X-ray analysis within this framework. During training, carefully designed system prompts guide the agent to follow a structured "think-then-answer" reasoning pattern, where the model first explicitly articulates its step-by-step reasoning (enclosed in delimiter tags), followed by presenting final disease probability predictions inside a box block. The GRPO algorithm optimizes the model by providing reward signals that incentivize both accurate predictions and strict adherence to the specified output format. Training is conducted for up to 3 epochs on 8 A100 GPUs, with a batch size of 512 and a learning rate of 1.0e-6. The evaluation framework assesses performance by checking for format adherence using regular expression pattern matching and by comparing disease probability predictions against ground truth labels. 

{
	\small
	\bibliographystyle{IEEEtran}
	\bibliography{refs}
}

\clearpage


\vspace{0.75cm}

\setcounter{figure}{0}
\setcounter{table}{0}
\renewcommand{\figurename}{Supplementary Figure}
\renewcommand{\tablename}{Supplementary Table}

\end{document}